\documentclass[iicol]{sn-jnl}

\usepackage{natbib}
\usepackage{graphicx}%
\usepackage{multirow}%
\usepackage{amsmath,amssymb,amsfonts}%
\usepackage{amsthm}%
\usepackage{mathrsfs}%
\usepackage[title]{appendix}%
\usepackage{xcolor}%
\usepackage{textcomp}%
\usepackage{manyfoot}%
\usepackage{booktabs}%
\usepackage{algorithm}%
\usepackage{algorithmicx}%
\usepackage{algpseudocode}%
\usepackage{listings}%
\usepackage{bm}
\usepackage{amsmath}
\usepackage{amssymb}

\newcommand{\model}{Cosh-DiT}

\theoremstyle{thmstyleone}%
\theoremstyle{thmstyletwo}%

\theoremstyle{thmstylethree}%

\begin{document}

\title[Article Title]{Cosh-DiT: Co-Speech Gesture Video Synthesis via Hybrid Audio-Visual Diffusion Transformers}


\author*[1,3]{\fnm{Yasheng} \sur{Sun}}\email{sunyasheng01@yeah.net}
\equalcont{These authors contributed equally to this work.}

\author[2]{\fnm{Zhiliang} \sur{Xu}}\email{xuzhiliang@baidu.com}
\equalcont{These authors contributed equally to this work.}

\author[2]{\fnm{Hang} \sur{Zhou}}\email{zhouhang09@baidu.com}

\author[4]{\fnm{Jiazhi} \sur{Guan}}\email{guanjz20@mails.tsinghua.edu.cn}

\author[5]{\fnm{Quanwei} \sur{Yang}}\email{yangquanwei@mail.ustc.edu.cn}

\author[2]{\fnm{Kaisiyuan} \sur{Wang}}\email{wangkaisiyuan@baidu.com}

\author[2]{\fnm{Borong} \sur{Liang}}\email{liangborong@baidu.com}

\author[2]{\fnm{Yingying} \sur{Li}}\email{liyingying05@baidu.com}

\author[2]{\fnm{Haocheng} \sur{Feng}}\email{fenghaocheng@baidu.com}

\author[2]{\fnm{Jingdong} \sur{Wang}}\email{wangjingdong@outlook.com}

\author[6]{\fnm{Ziwei} \sur{Liu}}\email{ziwei.liu@ntu.edu.sg}

\author[3]{\fnm{Koike} \sur{Hideki}}\email{koike@c.titech.ac.jp}

\affil*[1]{\orgdiv{Center of Excellence for Generative AI}, \orgname{King Abdullah University of Science and Technology}, \orgaddress{\street{Thuwal}, \postcode{23955-6900}, \state{Jeddah}, \country{Saudi Arabia}}}

\affil[2]{\orgdiv{Baidu VIS}, \orgaddress{\street{Shangdi 10th Street},  \postcode{100085}, \city{Beijing}, \country{China}}}

\affil[3]{\orgdiv{School of Computing}, \orgname{Tokyo Institute of Technology}, \orgaddress{\street{Ookayama, Meguro-ku}, \postcode{152-8550}, \state{Tokyo}, \country{Japan}}}

\affil[4]{\orgdiv{Department of Computer Science and Technology}, \orgname{Tsinghua University}, \orgaddress{\street{Haidian District, Shuangqing Road}, \postcode{100084}, \city{Beijing}, \country{China}}}

\affil[5]{\orgdiv{Department of Electronic Engineering and Information Science}, \orgname{University of Science and Technology of China}, \orgaddress{\street{Baohe District, JinZhai Road}, \postcode{230026}, \city{Hefei}, \state{Anhui}, \country{China}}}

\affil[6]{\orgdiv{S-Lab}, \orgname{Nanyang Technological University}, \orgaddress{\street{50 Nanyang Avenue},  \postcode{639798}, \country{Singapore}}}



\abstract{Co-speech gesture video synthesis is a challenging task that requires both probabilistic modeling of human gestures and the synthesis of realistic images that align with the rhythmic nuances of speech. To address these challenges, we propose \textbf{\model{}}, a \textbf{Co-s}peech gesture video system with \textbf{h}ybrid \textbf{Di}ffusion \textbf{T}ransformers that perform audio-to-motion and motion-to-video synthesis using discrete and continuous diffusion modeling, respectively.  
First, we introduce an \textbf{audio Diffusion Transformer (Cosh-DiT-A)} to synthesize expressive gesture dynamics synchronized with speech rhythms. To capture upper body, facial, and hand movement priors, we employ vector-quantized variational autoencoders (VQ-VAEs) to jointly learn their dependencies within a discrete latent space. Then, for realistic video synthesis conditioned on the generated speech-driven motion, we design a \textbf{visual Diffusion Transformer (Cosh-DiT-V)} that effectively integrates spatial and temporal contexts. Extensive experiments demonstrate that our framework consistently generates lifelike videos with expressive facial expressions and natural, smooth gestures that align seamlessly with speech.}




\maketitle

\section{Introduction}\label{intro}

Driving a portrait with audio is of significant importance in a variety of applications~\cite{seeger2021texting,pizzi2023chatbot,adam2021ai,roesler2021meta,seitz2022can}, such as digital human creation, human-computer interaction, and live-streaming e-commerce. 
Many works~\cite{zhou2021pose,thies2020neural,li2021write,park2022synctalkface,cheng2022videoretalking,ye2023geneface,fan2022faceformer} on this domain focus on facial animation around head region, which limits their applications. It is essential to synthesize gesture movements that conform to human speech to present a comprehensive virtual human solution. For the task of co-speech gesture synthesis, most studies~\cite{ahuja2020no,ahuja2020style,li2021audio2gestures,liu2022learning,ao2023gesturediffuclip,zhu2023taming} only target on intermediate 3D motion representations such as 3D
key points and 3D parametric models~\cite{SMPL-X:2019,SMPL:2015}. 
Despite advancing AR/VR and gaming~\cite{fu2022systematic,farouk2022studying}, these approaches are less suited for tasks like virtual anchor creation~\cite{huang2024make}, where realistic videos provide a more vivid and immersive experience. 


However, previous co-speech gesture studies cannot be easily transformed into the real-world domain.
Specifically, videos with co-speech gestures normally record only the upper body of a target person, making 3D reconstruction methods~\cite{SMPL-X:2019,cai2023smplerx} unstable. Particularly when the target is moving, without capturing the legs and feet, 3D reconstruction systems would find it hard to render reasonable results~\cite{yi2023generating}. On the other hand, leveraging 2D or 3D key points cannot provide enough details for human hands and expression. As a result, previous gesture representations are not suitable off-the-shelf preliminary tools for the co-speech gesture video synthesis task. While researchers also explored directly generating co-speech gesture videos~\cite{liu2022audio,he2024co}, their visual generator is not sufficient for producing realistic results. Very recently, VLOGGER~\cite{corona2024vlogger} leverages a two-stage system and uses 3D representations, but they only produce results with subtle movements. The task of achieving realistic co-speech gesture video synthesis is still under-explored and remains an open problem.



\begin{figure}[t]
  \includegraphics[width=0.475\textwidth]{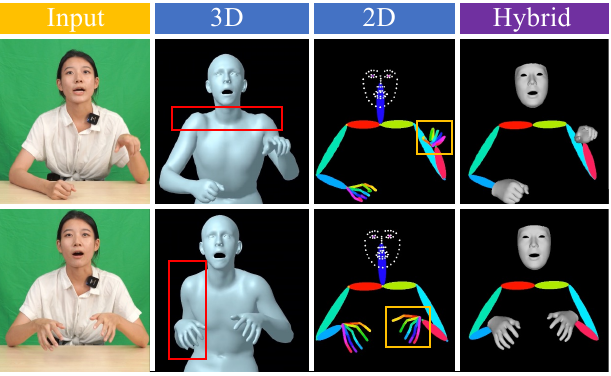}
  \caption{\textbf{Reconstruction with Varied Representations.} Holistic 3D reconstruction via SMPL-X often exhibits inaccuracies in limb and joint positioning (\textcolor{red}{red} box). While 2D pose estimation (e.g., DwPose) provides accurate body joint localization, it struggles to represent complex hand gestures effectively (\textcolor{yellow}{yellow} box). Combination of them leads to precise overly.} 
  \label{fig:hybrid_recon}
\end{figure}

\begin{figure*}
  \includegraphics[width=\textwidth]{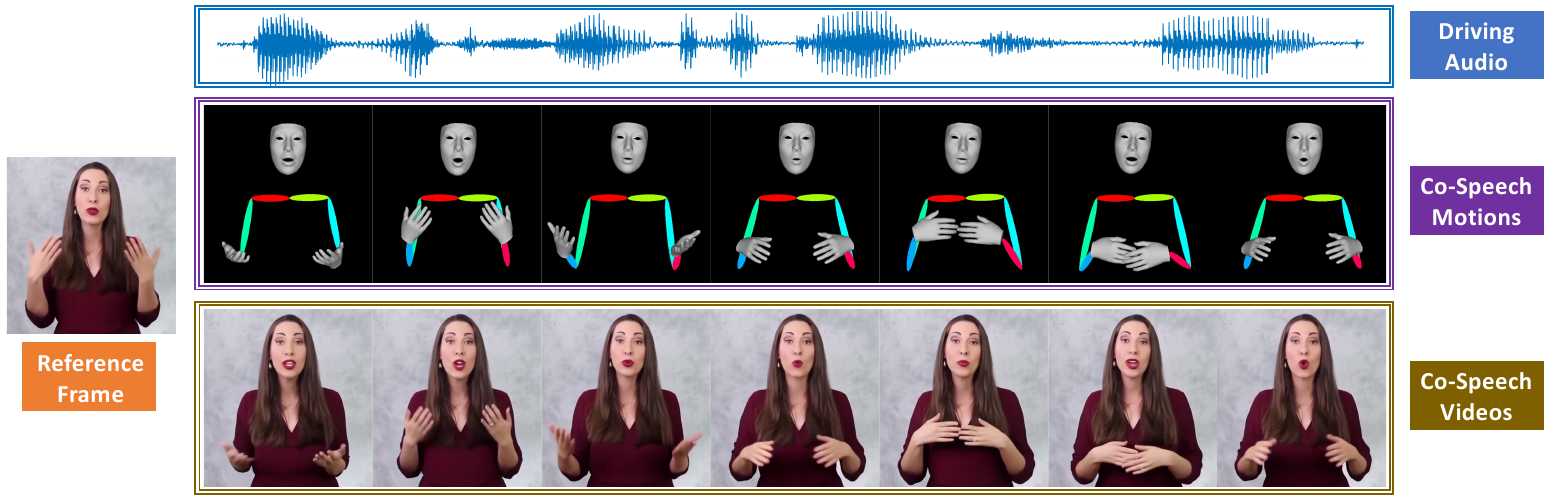}
  \caption{\textbf{Illustration of \model{} System.} Given a human speech input and an arbitrary reference image, our framework consistently generates lifelike videos with synchronized gestures. The synthesized videos feature natural, rhythmic movements and expressive facial and hand gestures, capturing vivid details for a realistic portrayal.}
  \label{fig:teaser}
\end{figure*}

To address this problem, we propose \textbf{\model{}}, a system illustrated in Fig.~\ref{fig:teaser}, which synthesizes realistic \textbf{Co-s}peech gesture videos using \textbf{h}ybrid audio-visual \textbf{Di}ffusion \textbf{T}ransformers. Our insight is to \emph{identify a suitable intermediate motion representation, on top of which audio-visual Diffusion Transformers are designed for realistic video synthesis}.
We argue that such representation should precisely deliver human motions and can be easily employed by video generation models. As a result, we first focus more on whether the representation can be well-overlayed in videos as shown in Fig.~\ref{fig:hybrid_recon}.
Specifically, leveraging parametric hand models~\cite{pavlakos2024reconstructing} not only conveys the depth difference and occlusion information but also effortlessly constrains finger shape and length. In contrast, the upper body movements can be well depicted through 2D skeletons. Additionally, both the Mano hand and body poses can be precisely and robustly detected by off-the-shelf models~\cite{pavlakos2024reconstructing,yang2023effective}, facilitating the usage of a large number of raw videos as high-quality training data. 

To learn the coordination of such mixed representations, we propose a discrete audio-driven Diffusion Transformer, \textbf{Cosh-DiT-A}, for motion synthesis. It takes a novel path that operates directly on encoded token indices. This approach differs from prior latent motion diffusion methods~\cite{chen2023executing} by offering a more abstract state space and enabling stronger motion priors~\cite{gu2022vector}. Another advantage is that such diffusion-based strategies iteratively take into account the whole temporal sequence, naturally avoiding issues like uni-directional bias and cumulative prediction errors of step-by-step generation from previous studies~\cite{yi2023generating,jiang2023motiongpt}. However, modeling such hybrid motion representation is non-trivial, especially connecting the 2D upper-body poses and 3D hand parametric models. 
Thus, we first extend the upper-body poses with the projected key points of hand motions, both of which are encoded into tokens and jointly modeled. Then a Geometric-Aware Alignment Module is carefully designed to optimize the hand position on the fly by projection restriction, seamlessly ensuring the holistic movements of hybrid motion representations. 

Synthesizing realistic video frames conforming to pose motions requires not only exploiting the appearance information of speaker identity but also spatially aligning with the temporal movements. 
Although extending the UNet diffusion backbones~\cite{blattmann2023stable,zhang2024mimicmotion,corona2024vlogger} with temporal attention has shown its effectiveness on realistic video synthesis, such approaches still struggle on detail synthesis such as hand or facial parts, which severely affects the visual aesthetics. 
In our work, we take advantage of recent advances in Diffusion Transformers (DiTs)~\cite{peebles2023scalable,esser2024scaling} for video generation~\cite{gao2024lumin-t2x,yang2024cogvideox}. A visual Diffusion Transformer network, \textbf{Cosh-DiT-V}, is devised to jointly model temporal-spatial association, providing higher model capacity and thus achieving improvement in terms of image quality. 

Our contributions are summarized as follows: 
\begin{itemize}
    \item We propose a two-stage DiT-based system, \textbf{Cosh-DiT}, to achieve realistic audio-driven co-speech gesture video synthesis, with the hydration of discrete and real-world domain diffusion transformers. 
    \item A vector-quantized audio-visual diffusion transformer is proposed to model the joint probability of human speech and the hybrid form of gesture representation, which is further enhanced by a \emph{Geometric-Aware Alignment Module}. 
    \item We delicately design a novel human motion diffusion transformer to extensively exploit the spatial-temporal motion and appearance features without the loss of speaker texture details.
\end{itemize}

\section{Related Work}\label{related}
\subsection{Co-Speech Gesture Generation.} 
Co-speech gesture generation focuses on synthesizing sequences of speech-synchronized gestures for talking portraits. Early work in this field relied on rule-based approaches~\cite{cassell2001beat,kopp2004synthesizing,levine2010gesture,poggi2005believable,marsella2013virtual,cassell1994animated,huang2012robot} to map human speech to pre-defined motion templates. Although effective, this method is both time-intensive and labor-intensive, requiring expert knowledge to design the rules and create motion templates. To overcome these limitations, later research introduced data-driven approaches~\cite{ferstl2018investigating,kucherenko2019analyzing,li2021audio2gestures,yoon2020speech,liu2022learning,qi2024emotiongesture,yin2023emog}, leveraging deep learning to model the joint probability of speech and human motion. Various architectures, including CNNs~\cite{habibie2021learning}, RNNs~\cite{yoon2019robots}, Transformers~\cite{bhattacharya2021text2gestures}, and VQ-VAEs~\cite{liu2022audio,yazdian2022gesture2vec}, have been explored to improve performance.
To mitigate mode collapse in GAN-based approaches~\cite{goodfellow2020generative,ginosar2019learning,liu2022learning,qian2021speech,yoon2020speech}, recent works have adopted diffusion models~\cite{ao2023gesturediffuclip,zhu2023taming}, which enable high-fidelity gesture synthesis. For more lifelike results, some approaches aim to synthesize holistic body movement~\cite{yi2023generating,habibie2021learning,liu2024towards}, incorporating both gestures and facial expressions. However, these methods often target to animate intermediate 3D representations, such as SMLP-X~\cite{pavlakos2019expressive}, typically reconstructed from video data and prone to reconstruction errors. On the other hand, real-world applications often require direct video synthesis for smoother and more reliable animations.

\subsection{Human Motion Video Synthesis.} 
With advances in text-to-image diffusion models~\cite{dhariwal2021diffusion}, recent studies~\cite{villegas2022phenaki,xing2024make,khachatryan2023text2video,ho2022imagen,blattmann2023stable,bar2024lumiere,he2022latent,guo2023animatediff} have extended the diffusion framework to video synthesis, significantly improving video quality, particularly in detail synthesis. Human motion animation~\cite{hu2024animate,liu2022disco,xu2024magicanimate,zhou2022magicvideo,karras2023dreampose,zhang2024mimicmotion} focuses on synthesizing natural body sequences conditioned on various control signals, such as dense pose~\cite{guler2018densepose}, skeleton~\cite{hu2024animate}, and mesh~\cite{corona2024vlogger}. Spatial control signals, like pose sequences, usually offer precise alignment, whereas audio signals convey complex linguistic and rhythmic information, making the task more challenging.
To animate videos from human speech, a common approach is to first transform audio into motion representations and then use these motion templates to animate a speaker’s appearance in a two-stage framework~\cite{liu2022disco,yang2023diffusestylegesture,pang2023bodyformer,nyatsanga2023comprehensive,he2024co}. However, many previous approaches~\cite{ginosar2019learning,qian2021speech} rely on GAN-based video synthesis backbones, which often introduce unrealistic artifacts.
To enhance visual quality, recent work in audio-driven motion animation has employed UNet-based diffusion models~\cite{lin2024cyberhost,corona2024vlogger}. 
However, generating speech-synchronized upper-body movements with natural facial expressions and hand gestures remains an unsolved challenge. Our work explores a hybrid skeleton-mesh intermediate representation as motion guidance to tackle this issue and improve synchronization and realism in video synthesis.

\begin{figure*}[t]
    \centering
    \includegraphics[width=1.0\linewidth]{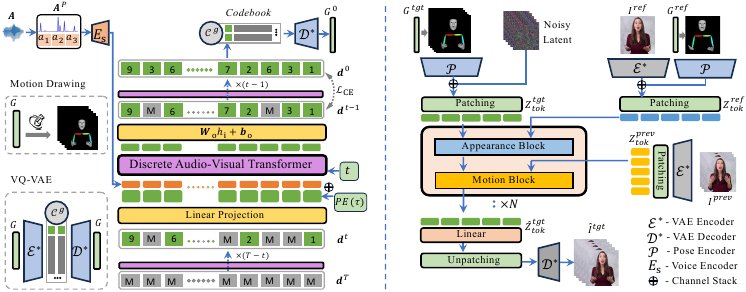}
    \caption{\textbf{Overview of \model{} System.} On the left is the Cosh-DiT-A model which takes audio signals as input and processes discrete gesture representations. The DiT model takes masked noisy tokens as input and predicts the quantized gesture tokens. On the right is the Cosh-DiT-V model for co-speech gesture video synthesis. It takes rendered gesture representations, appearance references and precious motion frames as input to produce video frames.
}
\label{fig:pipeline}

\end{figure*}

\subsection{Discrete Representation on Generative Model.}
To fully exploit the visual prior on visual generative models, researchers have proposed to encode the raw pixels to a codebook~\cite{esser2021taming,rombach2022high,zhou2022codeformer} by vector quantized variational autoencoders (VQ-VAEs).
Recent studies~\cite{liu2024towards,yi2023generating,jiang2023motiongpt} have shown that discrete representations can significantly enhance the quality of human motion synthesis. However, these approaches often rely on auto-regressive strategies~\cite{yi2023generating,jiang2023motiongpt} that generate motion step by step, leading to challenges like uni-directional bias and cumulative prediction errors. To address these issues, some research~\cite{gu2022vector} in text-to-image generation has applied diffusion techniques to discrete code generation where the de-noising strategy is replaced with discrete state transition via a transition matrix. But this approach remains largely unexplored in the realm of human motion synthesis, which might have potential on effective human motion prior modeling.

\section{Methodology}\label{method}

We present the \textbf{\model{} System}, designed to generate lifelike videos with facial and body movements synchronized with audio input. The overall pipeline is illustrated in Fig.~\ref{fig:pipeline}, following a two-stage process of audio-to-motion and motion-to-video synthesis. We detail the first stage, where we leverage a hybrid motion representation constrained within a well-defined space in Sec.~\ref{sec:3.1}. Following this, we outline the architecture for co-speech gesture video synthesis, using the predicted motion representations (Sec.~\ref{sec:3.2}).

\subsection{Audio-Driven Gesture Synthesis}
\label{sec:3.1}

The first stage targets to recover human gestures representations \( G = (G_{(1)}, G_{(2)}, \dots, G_{(F)}) \) from their corresponding speech representations \( A = (A_1, A_2, \dots, A_F) \) with \( F \) as the sequence length. 
In contrast to previous studies that solely use 2D~\cite{NEURIPS2019_7ca57a9f} or 3D~\cite{liu2024towards} human body representations, our framework explores a hybrid motion representation to precisely describe movements. As the hybrid representation lacks consistent semantic meanings,
to model such flexible representations, we introduce discrete diffusion strategies to exploit human motion priors and geometric-aware alignment to explicitly restrict plausible actions. The whole learning process is performed on a vector-quantized version of our gesture representations, through the discrete Audio-driven Co-speech Gesture Diffusion Transformer, namely the \textbf{Cosh-DiT-A}. 
 
\paragraph{Hybrid Gesture Representation.} We represent the upper body using 2D poses, which can be accurately detected by pre-trained networks like DWPose~\cite{yang2023effective}. Specifically, a total of 17 key points, capturing both arm and palm movements, are incorporated into the first-stage learning.  To ensure consistent expression in upper body 2D pose representation, rather than using raw 2D coordinates, we convert these points into angles and lengths, with angles further represented as 6D representation. Thus the whole 2D
poses are represented by a vector of $16 \times 6 + 16 + 2 = 114$ length. To incorporate geometric priors for hand movements, we adopt the MANO parametric hand model~\cite{MANO:SIGGRAPHASIA:2017} detected by~\cite{pavlakos2024reconstructing}. We utilize rotation matrices of each hand's 16 joints, converted into a 6D representation~\cite{zhou2019continuity} for consistency and usability. The hands' representations are represented as a vector of $16 \times  6 \times 2 = 192$ dimensions. 
Thus the final hybrid gesture is formulated as a vector $G \in \mathbb{R}^{306}$.  This unified approach enables a cohesive and comprehensive representation of upper body and hand movements. 



\begin{figure}[t]
    \centering
    \includegraphics[width=0.6\linewidth]{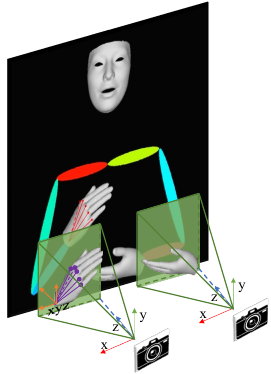}
    \caption{
    \textbf{Geometric-Aware Alignment}. This module ensures that the key points of the hand model \( [X_{k}, Y_{k}, Z_{k}]^\top \) in camera space are accurately projected to the 2D points \( [u_k, v_k]^\top \) on the image plane by optimizing the translation vector \( \mathbf{xyz} = [X_{tran}, Y_{tran}, Z_{tran}]^\top \). 
}
\label{fig:proj_align}

\end{figure}

\paragraph{Geometric-Aware Alignment.}
To composite 3D hand gestures and 2D body poses, the 3D hand points from camera space must be projected to align with their corresponding 2D points in the image plane. This is achieved by optimizing the camera projection parameters as shown in Fig.~\ref{fig:proj_align}. 
After perspective projection, the 3D coordinates \( \bm{k_k} = (X_k, Y_k, Z_k)^\top \) of the palm joints should align with the 2D coordinates \( \bm{K_K} = (u_k, v_k)^\top \) in the image plane. Formally, this restriction is described by the following equation:


\begin{equation}
\begin{bmatrix}
u_k \\
v_k \\
\end{bmatrix}
=
\frac{1}{Z_k + Z_{tran}}
\begin{bmatrix}
f_x & 0 & c_x \\
0 & f_y & c_y \\
0 & 0 & 1 \\
\end{bmatrix}
\begin{bmatrix}
X_k + X_{tran}\\
Y_k + Y_{tran}\\
Z_k + Z_{tran}\\
\end{bmatrix}   . 
\end{equation}

Here, \( (X_k, Y_k, Z_k) \) are the 3D coordinates of the palm joint in the camera space, \( (u_k, v_k) \) are the 2D coordinates of the palm joint in the image, \( f_x \) and \( f_y \) are the focal lengths of the camera, and \( c_x \) and \( c_y \) are the coordinates of the principal point of the camera. The depth \( Z_k \) is the distance from the camera to the point along the \( Z \)-axis. 
The translation parameters \( \mathbf{xyz} = (X_{tran}, Y_{tran}, Z_{tran})^\top \) represent the unknown offsets of the hand, which we iteratively optimize to ensure that the 3D points align with their corresponding 2D locations after projection. 
The optimization objective for video frame \( f \) is formulated as:
\[
\mathcal{L}_{proj}^f = (\hat{u}_k^f - u_k^f)^2 + (\hat{v}_k^f - v_k^f)^2 + \lambda_{tr} (\hat{Z}_{tran}^f - \hat{Z}_{tran}^{f-1})^2.
\]
Since \( Z_{tran} \) influences the scale of the human hand, we enforce temporal continuity on the depth translation \( \hat{Z}_{tran} \) to ensure smoothness in the predicted hand motion. Here, \( \lambda_{tr} \) is the weight coefficient that controls the regularization.

\paragraph{Discrete Diffusion Modeling.}
Given a vector of human motion descriptors \( G_{\{1,\cdots,F\}} \in \mathbb{R}^{F \times 306} \), where \( F \) represents the sequence length, the VQ-VAE encoder \( \bm{\mathcal{E}}^{*} \) compresses it with a temporal resolution reduction of \( \times 8 \), producing a latent feature \( \bm{z}_g \in \mathbb{R}^{\frac{F}{8} \times 128} \). Here, 128 denotes the feature dimension of the codebook \( {\mathcal{C}}^g \).
The closest code \( \bm{z}_q \) is then identified from the codebook \( {\mathcal{C}}^{g} \) based on the quantization process, formally
\begin{equation}
	\bm{z}_q =  \operatorname{arg\,min}_{\bm{z}_c \in \mathcal{C}} \| \bm{z}_c - \bm{z}_g \|_2^2.
\end{equation}
The index of \( \bm{z}_q \) is indicated as \( \bm{d} \in \mathbb{R}^{\frac{F}{8}} \). 
Therefore, the gestures can be represented by a discrete vector with the same length.
In contrast to typical diffusion model that iteratively denoises Gaussian noise to obtain the target \(x_0\), the discrete diffusion directly operates on the constructed discrete vector $\bm{d}^t$. Here $t$ is the denoising step. 


Particularly, we define the probability of transition from \(\bm{d}^{t-1}\) to \(\bm{d}^t\) using matrices \( [\bm{Q}^t]_{mn} = q(\bm{d}^t_f = m | \bm{d}^{t-1}_f = n) \in \mathbb{R}^{K \times K} \), where the $f$ indicates $f$th element of $\bm{d}$ and $K$ is the codebook size. The \emph{forward} Markov diffusion process for the sequence of tokens is then written as:
\begin{equation}
\vspace{-0.2cm}
q(\bm{d}^t_f|\bm{d}^{t-1}_f) = \bm{v}^\top(\bm{d}^{t}_f) \bm{Q}^t \bm{v}(\bm{d}^{t-1}_f),
\label{eq:markov}
\vspace{-0.02cm}
\end{equation}
where \( \bm{d}^{t}_f \) is a one-hot column vector of length \( K \) with a 1 at position \( \bm{d}^{t}_f \). The categorical distribution over \( \bm{d}^{t}_f \) is represented by the vector \( \bm{Q}^t \bm{v}(\bm{d}^{t-1}_f) \). The posterior for this diffusion process is expressed as:
\begin{equation}
\vspace{-0.1cm}
\begin{split}
q(\bm{d}^{t-1}_f|\bm{d}^t_f, \bm{d}^0_f) = \frac{q(\bm{d}^t_f|\bm{d}^{t-1}_f, \bm{d}^0_f) q(\bm{d}^{t-1}_f|\bm{d}^0_f)}{q(\bm{d}^t_f|\bm{d}^0_f)} \\ 
= \frac{
	\left(\bm{v}^\top(\bm{d}^{t}_f) {\bm{Q}}^t \bm{v}(\bm{d}^{t-1}_f)\right) 
	\left(\bm{v}^\top(\bm{d}^{t-1}_f) \overline{\bm{Q}}^{t-1} \bm{v}(\bm{d}^0_f)\right)}
	{\bm{v}^\top(\bm{d}^{t}_f) \overline{\bm{Q}}^t \bm{v}(\bm{d}^0_f)},
\end{split}
\vspace{-0.1cm}
\end{equation}
where \( \overline{\bm{Q}}^t = \bm{Q}^t \cdots \bm{Q}^1 \). 

Starting from the noise \(\bm{d}^T\) composed of masked tokens $\textbf{M}$, the \emph{reverse} process denoises the latent variables gradually to recover the original data \( \bm{d}^0 \) by sequential sampling from the reverse distribution \( q(\bm{d}^{t-1}_f|\bm{d}^t_f, \bm{d}^0_f) \). Since \( \bm{d}^0 \) is unknown during inference, we train a Transformer network to approximate the transition distribution \( p_\theta(\bm{d}^{t-1}|\bm{d}^t) \), which relies on the entire data distribution.

\paragraph{Audio-Driven Discrete Diffusion Transformer.}

Using the discrete representation of human gestures, our discrete audio-visual diffusion transformer Cosh-DiT-A is designed to model the association between audio and gestures. Given an input of human speech, the model synthesizes motions that are rhythmically consistent with the audio. To enhance rhythm processing, we preprocess the speech into BEAT features \( \bm{A}^B = (A_1^B, A_2^B, \dots, A_F^B) \) as described in~\cite{zhuang2022music2dance}. Additionally, we employ the Wav2Vec2 model~\cite{baevski2020wav2vec} to extract semantic content \( \bm{A}^C = (A_1^C, A_2^C, \dots, A_F^C) \) from the audio. They are concatenated as \(\bm{A}^P = (\bm{A}^B, \bm{A}^C)\) to formulate the input of our voice network $\bm{E}_s$. 

The model is based on a Transformer architecture~\cite{vaswani2017attention}, with attention mechanisms applied across the temporal dimension to effectively capture temporal dependencies.
Position encoding, denoted as \( \text{PE}(\tau) \), is added along the temporal axis, and a diffusion time step \( t \) is introduced to modulate the convolution operations, allowing the network to remain time-aware. Formally, given an input sequence of audio features \( \bm{A}^P \) and noisy motion tokens \( \bm{d}^{t} = (d_1^{t}, d_2^{t}, \dots, d_F^{t}) \), they are concatenated and fed into the transformer layers to recover the \(\bm{d}^0\). The final layer applies a Softmax function to yield the probability distribution \( p(\bm{d}^{t-1} | \bm{d}^{t}, \bm{A}^P ) \) at each time step:
\begin{align}
p(\bm{d}^{t-1} | \bm{d}^{t}, \bm{A}^P ) = \text{Softmax}(\bm{W}_o h_i + \bm{b}_o)    
\end{align}

where \( h_i \) represents the hidden state at position \( i \), and \( \bm{W}_o \) and \( \bm{b}_o \) are learned parameters. 
Then \(\hat{\bm{d}}^0\) is calculated following the similar protocol~\cite{gu2022vector} of re-parametrization technique from \( \bm{d}^{t-1}\), on top of which a Cross-Entropy~\cite{zhang2018generalized} loss \( \mathcal{L}_{CE} (\hat{\bm{d}}^0, \bm{d}^0) \) is applied.


\subsection{Co-Speech Gesture Video Synthesis}
\label{sec:3.2}

This stage aims to animate reference portrait following the human motions obtained above.
In contrast to UNet-based networks~\cite{hu2024animate}, which capture temporal context through attention aggregation along the channel axis, we directly flatten all video patches within a sliding window for effective spatial-temporal modeling.
It is achieved via a visual Co-speech Gesture Diffusion Transformer, namely the \textbf{Cosh-DiT-V}, as illustrated on the right side of Fig.~\ref{fig:pipeline}. 
Similar to the first stage, both the training and inference processes in this stage adhere to the diffusion paradigm. For simplicity, the time step \( t \) notation is omitted.

\begin{figure*}[t]
    \centering
    \includegraphics[width=1.0\linewidth]{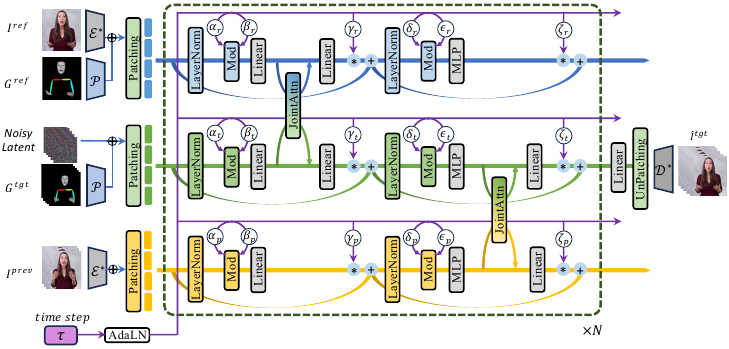}
    \caption{\textbf{Appearance Motion (AM) - DiT Architecture.} Three types of information are simultaneously fed into the AM-DiT block. The noisy latents, conditioned on gesture guidance, are combined with the reference person identity information and previous motion features, which are iteratively integrated through a joint attention operation.
}
\label{fig:mmdit}

\end{figure*}

\paragraph{Video Patchifying.}
The raw video frames \( I_{\{1,\cdots,K\}} = \{ I_{(1)}, \cdots, I_{(K)} \} \) are first encoded into the VAE latent space for subsequent processing. Formally, this is expressed as:
\[
Z = \bm{\mathcal{E}}^{*}(I_{\{1,\cdots,K\}})
\]
We adopt the encoder \( \bm{\mathcal{E}}^{*} \) and decoder \( \bm{\mathcal{D}}^{*} \) architecture, following a similar protocol as in~\cite{yu2023language}. For a video clip with \( K \) frames, the first frame is encoded separately, while the subsequent frames are downsampled by a factor of 4 to balance the spatial and temporal information. Additionally, the spatial dimension is compressed by a factor of 8. Thus, for a video clip \( I_{\{1,\cdots,K\}} \in \mathbb{R}^{K \times H \times W \times 3} \), after encoding, the latent feature becomes:
\[
Z \in \mathbb{R}^{\frac{4+(K-1)}{4} \times \frac{H}{8} \times \frac{W}{8} \times 4}
\]
To recover the video frames in pixel space, the latent features \( Z' \) are passed through the VAE decoder, yielding the output \( I'_{\{1,\cdots,K\}} = \bm{\mathcal{D}}^{*}(Z') \). For processing the encoded features \( Z \) with a transformer block, they are divided into a series of spatial-temporal patches \( Z_{patch} \in \mathbb{R}^{\frac{4+(K-1)}{4} \times \frac{H}{8P} \times \frac{W}{8P} \times P \times P \times 4} \), where the channel depth is 4. The sequence of patches, each of size \( P \times P \times 4 \), is then embedded into \( \bm{Z}_{tok} \in \mathbb{R}^{\frac{4+(K-1)}{4} \times \frac{H}{8P} \times \frac{W}{8P} \times F} \), with a channel size of \( F \).

\paragraph{Diffusion Transformer in the Video Domain.}

Given a sliding window with \( K \) target motion parameters, \( G^{tgt} = \{G^{tgt}_{(1)}, \dots, G^{tgt}_{(K)}\} \), and a reference frame \( I^{ref} \), the network aims to generate animated video frames \( I^{tgt} = \{I^{tgt}_{(1)}, \dots, I^{tgt}_{(K)}\} \) with consistent and lifelike movements.
To encourage consistent representation, these motion parameters are rendered into video frames in the image space using a 3D mesh renderer~\cite{pavlakos2024reconstructing,ravi2020pytorch3d}.

As illustrated in Fig.~\ref{fig:pipeline}, the video frames are patchified into tokens. 
For target and reference motion frames $G^{tgt}, G^{ref}$, a pose encoder \( \mathcal{\bm{P}} \) is designed to extract their spatial features, aligning with the latent space of VAE encoder $\bm{\mathcal{E}}^{*}$. Once aligned, their representations are concatenated to the corresponding video frames to enhance pose awareness. They are projected and embedded into target and reference tokens \( \bm{Z}_{tok}^{tgt}, \bm{Z}_{tok}^{ref} \), preparing them for subsequent processing by the transformer layers. 
By treating the spatial and temporal dimensions from a unified perspective, this approach grants the network greater flexibility in capturing spatial-temporal associations.
Note that smooth transitions in both motion and texture are essential for video synthesis, we also incorporate the preceding \( M \) frames, \( I^{prev} = \{ I^{prev}_{(1)}, \dots, I^{prev}_{(M)} \} \), to guide continuity in animation. These frames are directly encoded by a VAE encoder \( \bm{\mathcal{E}}^{*} \) and converted into a sequence of tokens \( \bm{Z}_{tok}^{prev} \).
To effectively utilize tokens from different sources such as \( G^{tgt} \), \( I^{ref} \) and \( I^{prev} \), 
we carefully design a stack of DiT blocks to progressively incorporate reference and motion information. After passing through them, 
target motion tokens \( \bm{Z}_{tok}^{tgt} \) are modulated by a stack of DiT blocks to \( \hat{\bm{Z}}_{tok}^{tgt} \), which are decoded back into video frames, producing \( \hat{I}^{tgt} = \{\hat{I}^{tgt}_{(1)}, \dots, \hat{I}^{tgt}_{(K)}\} \).

\paragraph{Stacked Iterative DiT Blocks.} 



As shown in Fig.~\ref{fig:pipeline}, we stack \( N \) DiT blocks to incorporate the speaker's appearance from the reference image \( I^{ref} \) and the continuous movement features from the preceding frames \( I^{prev} \). Rather than processing both types of information in a single DiT block, we propose an iterative approach that separately addresses appearance and movement features within dedicated blocks, given their inherent differences. In this approach, the portrait appearance information is first integrated into the target motion flow to enhance texture synthesis through the \textbf{Appearance Block}. Once the reference features are applied at the corresponding spatial locations, the \textbf{Motion Block} compares these with features from previous frames, refining appearance details and promoting temporal consistency in the generated human motions. 


Specifically, three flows are simultaneously passed through a sequence of \(N\) transformer blocks as shown in Fig.~\ref{fig:mmdit}, where they interact via a joint attention operation. Specifically, for each flow, the elements are mapped into queries (\(Q\)), keys (\(K\)), and values (\(V\)). The interaction is achieved by concatenating these elements and applying the attention mechanism, which can be formalized as:
\begin{equation}
\begin{aligned}
\text{JointAttn}(Q, K, V) &= \\
\text{softmax}\left(
\frac{[\mathbf{Q}_1 \| \mathbf{Q}_2] [\mathbf{K}_1 \| \mathbf{K}_2]^\top}{\sqrt{d_k}}
\right) 
& \cdot [\mathbf{V}_1 \| \mathbf{V}_2].
\end{aligned}
\end{equation}
After such operation, the values are separated back to their original streams.
To make these modules time-aware, the time step \(\tau\) is used to modulate each flow. The obtained parameters \((\alpha, \beta)\) and \((\sigma, \epsilon)\) scale and shift the input \(x\). This operation is formally written as:
\begin{equation}
    \text{Mod}(x) = \alpha(\tau) \cdot x + \beta(\tau).
\end{equation}
Another parameters \(\gamma\) and \(\zeta\) are used to control the input as a gate, applying \(\gamma(\tau) \cdot x\).

\section{Experiments}\label{exp}
\subsection{Experimental Settings}


\paragraph{Datasets.}
We collect a diverse dataset of 200 hours of video footage featuring centralized 2D speaking avatars from various sources, comprising 921 subjects in total. The test set is split into two parts, each containing a varied selection of video clips and frame-audio pairs: 18 clips from 4 subjects sourced from the PATs~\cite{ahuja2020style} dataset, and 21 clips from 21 subjects in our own collected dataset. Each video or audio clip has a duration of 7 to 10 seconds. Notably, the evaluation is performed in a zero-shot setting with no test identities were seen during training.

\paragraph{Implementation Details.}
All the videos are processed at 25 fps, resized with a height of 768 pixels while maintaining the original aspect ratio, and the audio is pre-processed to 16kHz.

\noindent\textbf{Cosh-DiT-A.}
In the first stage, the constructed codebook contains 128 entries, making the maximum value of $\bm{d}^t$ equal to 127. The sliding window size $F$ for gestures $G^{tgt}$ is set to 128 (approximately 5 seconds), with a conditional prefix motion length of $p=16$. This stage employs a transformer with 24 layers, trained with a batch size of 128 and a learning rate of $1 \times 10^{-4}$. 

For driving lip movements, a transformer is designed to predict the vertex offsets of the face model, following a protocol similar to FaceFormer~\cite{fan2022faceformer}. 

For geometric alignment, we use the Adam optimizer~\cite{kingma2014adam} to estimate the translation vector of the 3D hand model. The depth prior weight $\lambda_{tr}$ is empirically set to 10. To ensure accurate initial translation, we perform 500 iterations for the first frame and 250 iterations for subsequent frames. 

The generated motion vectors $G^{tgt}$, comprising both hand model parameters and body pose locations from the audio-driven gesture network, are first rendered into the image space using a 3D mesh renderer~\cite{pavlakos2024reconstructing,ravi2020pytorch3d} for subsequent processing in the second-stage video synthesis.

\noindent\textbf{Cosh-DiT-V.}
For the video diffusion transformer, the sliding window size $K$ is set to 25, and the motion frame count $M$ is 5. This stage is trained with a batch size of 8 and a learning rate of $1 \times 10^{-5}$. During training, one reference frame is randomly selected within a 30-frame range. Additionally, motion frames are randomly dropped with a probability of $\beta=0.2$.

\setlength{\tabcolsep}{5pt}
\begin{table*}[t] 
\begin{center}  \caption{\textbf{Quantitative results}. For FID, FVD and LPIPS the lower the better, and the higher the better for other metrics.}
\label{tab:quanti}
\begin{tabular}{lcccccccccc}
\toprule
 & \multicolumn{4}{c}{Static Assessment}& \multicolumn{4}{c}{Dynamic Assessment} \\
\cmidrule(lr){2-5} \cmidrule(lr){6-9}
Method & SSIM$\uparrow$  & FID$\downarrow$  & LPIPS$\downarrow$ & IDSim$\uparrow$  & FVD$\downarrow$   & Conf-head$\uparrow$ & Conf-body$\uparrow$ & Conf-hand$\uparrow$ \\

\midrule  
TS-MM      & 0.5612     & 172.41    & 0.5093    & 0.1781      & 1715.64               & 0.9695        & 0.8999        & 0.8294                        \\
TS-CX       & 0.6350    & 101.52     & 0.3561    & 0.4299      & 1526.19                & 0.9837        & 0.9208        & 0.7738                       \\
PT-MM      & 0.5731     & 165.91    & 0.5075    & 0.2546      & 1599.02               & 0.9830        & 0.8915        & 0.8224                        \\
PT-CX       & 0.6076    & 93.56     & 0.3892    & 0.4696      & 1342.13                & 0.9858        & 0.9238        & 0.7431                       \\
EG-CX       & 0.6231    & 131.52     & 0.3961    & 0.2937      & 1622.67                & 0.9813        & 0.9166        & 0.8354                       \\
\textbf{Ours}        & \textbf{0.6729}    & \textbf{77.51}   & \textbf{0.3175}    & \textbf{0.5918}     & \textbf{1118.94}                 & \textbf{0.9858}        & \textbf{0.9343}        & \textbf{0.8937}                       \\
\hline
S2G-MDD  & 0.5865       & 112.24     & \textbf{0.3496}         & 0.5962        & 1201.72       & \textbf{0.9837}        & 0.9449        & 0.6812          \\
\textbf{Ours}            & \textbf{0.6814}       & \textbf{79.78}      & 0.3519         & \textbf{0.6069}        & \textbf{881.83}        & 0.9825        & \textbf{0.9471}        & \textbf{0.8833}          \\
\bottomrule
\end{tabular}
\end{center}

\end{table*}

\paragraph{Comparison Approaches.} 
To tackle the challenge of generating 3D holistic body motions, \textbf{TalkShow} (\textbf{TS})\cite{yi2023generating} employs a compositional VQ-VAE combined with a cross-conditional autoregressive model to synthesize coherent and realistic motions. In their subsequent work, \textbf{ProbTalk} (\textbf{PT})\cite{liu2024towards} introduces product quantization (PQ) to encode complex holistic motions and utilizes a non-autoregressive model to enhance the vividness of co-speech motions. 
Another work \textbf{EMAGE} (\textbf{EG})\cite{liu2024emage} involves a Masked Audio Gesture Transformer facilitating joint training on audio-to-gesture generation and masked gesture reconstruction.
In contrast, \textbf{S2G-MDD}\cite{he2024co} 
focuses on synthesizing human motions from speech in the video domain. \textbf{S2G-MDD} employs a diffusion framework for latent motions but relies on a Generative Adversarial Network (GAN) for rendering, which results in lower video quality. 

\paragraph{Video Generation Paradigm.} 
We compare with state-of-the-art audio-driven body motion works, \textbf{TalkShow} (\textbf{TS})\cite{yi2023generating}, \textbf{ProbTalk} (\textbf{PT})\cite{liu2024towards} and \textbf{EMAGE} (\textbf{EG})\cite{liu2024emage}. However, these approaches only generate human motions in the form of SMPL-X~\cite{SMPL-X:2019}. To animate these motions within the video domain, we use widely adopted backbones, \textbf{ControlNeXt} (\textbf{CX})\cite{peng2024controlnext} and \textbf{MimicMotion} (\textbf{MM})\cite{zhang2024mimicmotion}. We combine these techniques into four configurations: \textbf{TS-CX}, \textbf{TS-MM}, \textbf{PT-CX}, \textbf{PT-MM} and \textbf{EG-CX}. 
Additionally, we incorporate co-speech gesture video works \textbf{S2G-MDD}\cite{he2024co} and compare on their evaluation datasets.

\paragraph{Evaluation Metrics.}
Our evaluation for co-speech video generation is conducted in terms of static and dynamic aspects, respectively. 
%
We first conduct static assessment on image quality metrics \textbf{SSIM}~\cite{wang2004image}, \textbf{LPIPS}~\cite{zhang2018unreasonable}, and \textbf{FID}~\cite{heusel2017gans}, which capture the structural, perceptual, and distributional aspects of image quality.
In addition, we conduct another assessment of identity similarity (\textbf{IDSim}) by calculating the cosine similarity between the generated frames and their used reference frames at the feature space of ArcFace~\cite{deng2019arcface}. 

In terms of dynamic assessment, we first evaluate the temporal coherence of the generated videos via \textbf{FVD}~\cite{unterthiner2018towards} following the video preparation instructions from DisCo~\cite{wang2023disco}.
Moreover, to evaluate the quality of the generated motion sequence, we use a 2D key-point detection approach~\cite{cao2017realtime} to assess the confidence scores in the three essential components, head, body, and hands, denoted as \textbf{Conf-head}, \textbf{Conf-body}, and \textbf{Conf-hand}, respectively.


\begin{figure*}[t]
    \centering
    \includegraphics[width=1.0\linewidth]{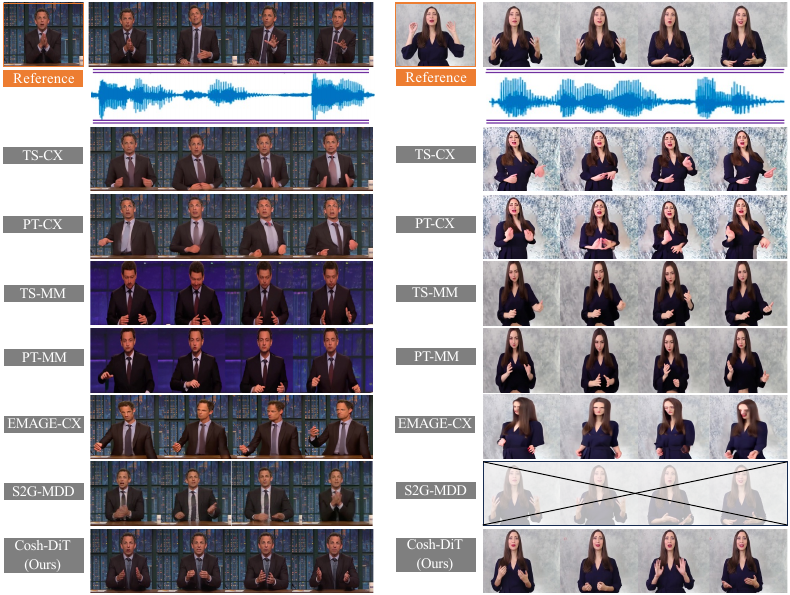}
    \caption{\textbf{Qualitative Results}. Our Cosh-DiT achieves the best visual quality among all its counterparts, which includes not only fine-grained textural details but also reasonable holistic human movements.
}
\label{fig:quali}

\end{figure*}

\subsection{Quantitative Evaluation}
Quantitative results are presented in Table~\ref{tab:quanti}, where our Cosh-DiT consistently outperforms all comparison methods across most evaluation metrics. When compared to the combined baselines, our approach demonstrates clear superiority in both image quality and identity preservation. Specifically, improvements in \textbf{SSIM}, \textbf{FID}, and \textbf{IDSim} highlight the enhanced structural accuracy of our generated results, particularly for dynamic components with significant motion. This indicates that our diffusion transformer-based pipeline, combined with the proposed hybrid representation, enables precise and high-fidelity per-frame synthesis. Moreover, when compared to existing co-speech gesture video works, S2G-MDD, our approach achieves superior performance, even outperforming their fine-tuned models on specific individuals.

In terms of dynamic performance, our Cosh-DiT achieves the lowest \textbf{FVD} score among all methods, suggesting superior temporal consistency in the synthesized video. Additionally, our method leads in keypoint-based metrics—namely, \textbf{Conf-body} and \textbf{Conf-hand}—demonstrating that our predicted body movements closely align with real human motion and are more reliable than those of other methods.

\begin{figure*}[t]
    \centering
    \includegraphics[width=1.0\linewidth]{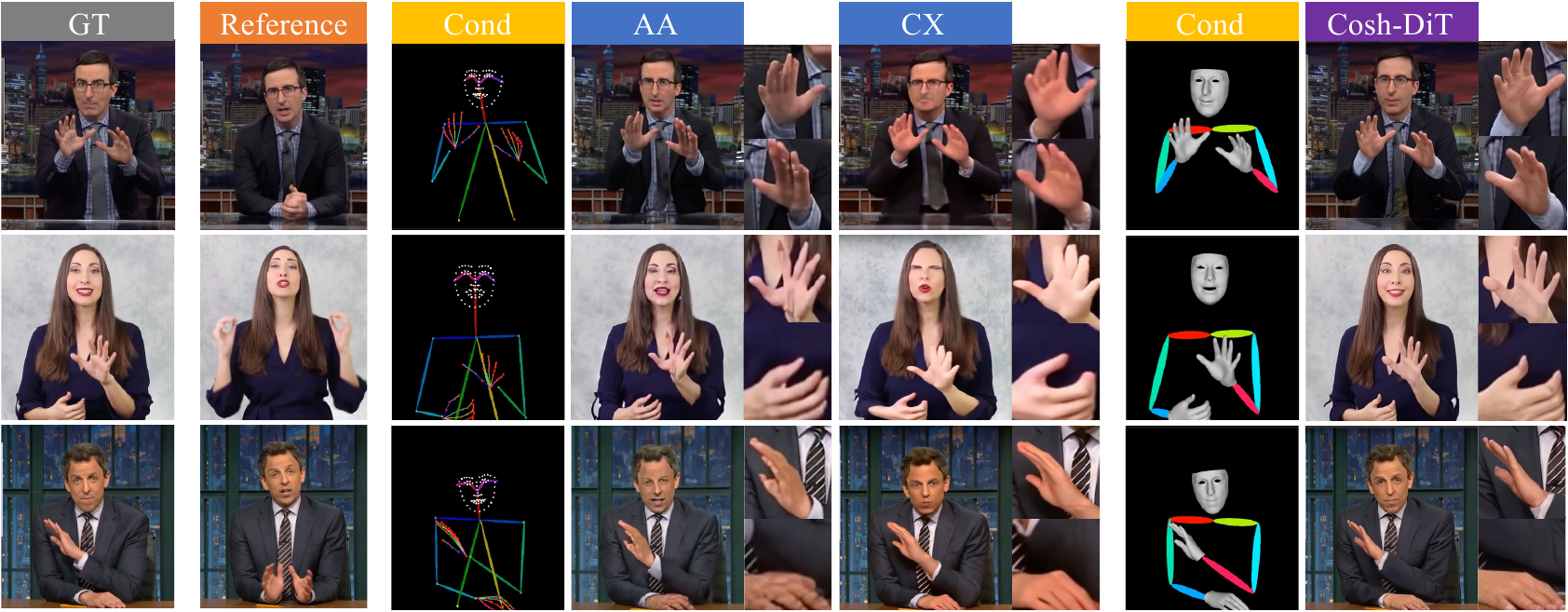}
    \caption{\textbf{Qualitative Comparison with State-of-the-Art Motion Animation Approaches}. We convert the motion format accordingly and compare our $\textbf{Cosh-DiT-V}$ with AnimateAnyone (AA)\cite{hu2024animate} and ControlNext (CX)\cite{peng2024controlnext}. The results highlight the superiority of our method in synthesizing realistic details around facial and hand regions.
}
\label{fig:hybrid_render}

\end{figure*}

\subsection{Qualitative Evaluation}
We also provide qualitative comparisons for better evaluation and the results are illustrated in Fig.~\ref{fig:quali}. We observe that when using \textbf{MimicMotion} (\textbf{MM}) as the video synthesis part, the generated results suffer from severe quality distortion and inconsistent facial identity. Similarly, \textbf{ControlNext} (\textbf{CX}) struggles with the quality distortion issue, it also fails to generate satisfactory hands and faces with clear structure and texture. In terms of audio-driven motion prediction, both \textbf{TalkShow} (\textbf{TS}) and \textbf{ProbTalk} (\textbf{PT}) produce rigid human body movements instead of physically reasonable articulated motion. \textbf{S2G-MDD}\cite{he2024co}  exhibit limitations in video quality, with noticeable artifacts, especially around the hand region.

Our \textbf{Cosh-DiT} achieves the best visual quality among all its counterparts, which includes not only fine-grained textural details but also reasonable holistic human movements.
Readers are recommended to refer to the supplementary materials for more details. 


\setlength{\tabcolsep}{7pt}
\begin{table}[t] 
\caption{\textbf{Quantitative Comparison with State-of-Art Motion Animation Approaches}. We compare our $\textbf{Cosh-DiT-V}$ with AnimateAnyone (AA)\cite{hu2024animate} and ControlNext (CX)\cite{peng2024controlnext}.}
\label{tab:quanti}
{
\renewcommand{\arraystretch}{1}
\begin{tabular}{lcccccccccc}
\toprule
Method & SSIM$\uparrow$  & FID$\downarrow$  & LPIPS$\downarrow$ & IDSim$\uparrow$  \\ 

\midrule  
AA      & 0.7655     & 42.64    & 0.1950    & 0.7881    
\\ 
CX       & 0.6504    & 55.64     & 0.3145    & 0.7191                  \\ 
\textbf{Cosh-DiT}     &   \textbf{0.8523}    & \textbf{24.62}   & \textbf{0.1477}    & \textbf{0.8938}        \\ 
\bottomrule
\end{tabular}
}
\label{tab:hybrid_render}

\end{table}

\subsection{Further Analysis} 

\paragraph{Discussion on Cosh-DiT-V Performance.}

To further validate the effectiveness of our proposed visual transformer, Cosh-DiT-V, we compare it against the state-of-the-art motion animator ControlNext (CX) \cite{peng2024controlnext}, as before. Additionally, we introduce AnimateAnyone (AA) as another baseline. Rather than conditioning on human speech, we use motion-specific inputs: pose skeletons for AA and CX, while our approach leverages a hybrid motion representation combining 2D body poses and 3D hand meshes.

As shown in Fig.~\ref{fig:hybrid_render}, our method achieves superior realism, particularly in rendering fine details around the hands. Quantitative results in Table~\ref{tab:hybrid_render} further confirm that our approach produces videos with visual quality comparable to the ground truth. 

\begin{figure}[t]
    \centering
    \includegraphics[width=1.0\linewidth]{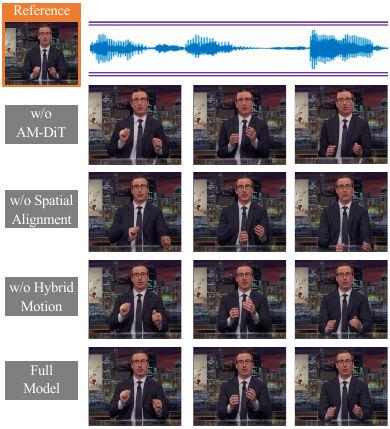}
    \caption{\textbf{Ablation study with visual results.}
    We demonstrate the results that hybrid motion, spatial alignment and AM-DiT blocks are separately removed.
}
\label{fig:ablation}

\vspace{-10pt}
\end{figure}

\setlength{\tabcolsep}{13.5pt}
\begin{table*}[t] 

\caption{\textbf{Ablation Study}. We conduct experiments by removing hybrid motion, spatial alignment and AM-DiT design.}
\label{tab:ablation}

\begin{center}  
\begin{tabular}{lcccccc} 
\toprule
Method & SSIM$\uparrow$  & FID$\downarrow$  & IDSim$\uparrow$  & Conf-body$\uparrow$ & Conf-hand$\uparrow$ \\   
\hline
w/o AM-DiT  &   0.6323  &  84.22   &   0.5742  &    0.9123     &   0.8793     \\
w/o Spatial Align   &  0.6419  &    89.24  &   0.5423  &     0.8723   &  0.8892          \\
w/o Hybrid Motion &  0.6584  &  79.34   &   0.5833  &   0.9155  &    0.8623  \\
Full Model   & \textbf{0.6729}    & \textbf{77.51}     & \textbf{0.5918}                & \textbf{0.9343}            &       \textbf{0.8937} \\
\bottomrule
\end{tabular}
\end{center}

\end{table*}

\setlength{\tabcolsep}{11pt}
\begin{table*}[t] 
\begin{center}

\caption{\textbf{User study Measured by Mean Opinion Scores.} Larger is higher, with the maximum value to be 5.}
\label{table:MOS}
\begin{tabular}{ccccccc}

\toprule

MOS on $\setminus$ Approach & PT-AA& TS-AA & PT-CX & TS-CX & S2G-MDD & \textbf{Ours} \\
\noalign{\smallskip}
\hline

Identity Similarity & 2.03 & 1.87 & 2.51 & 2.34 & 3.61 & \textbf{4.32} \\
Motion Naturalness  & 2.12 & 1.95 & 2.63 & 2.46 & 3.14 & \textbf{4.39} \\
Video Realness      & 1.94 & 1.78 & 2.43 & 2.25 & 3.42 & \textbf{4.62} \\
\bottomrule

\end{tabular}
\end{center}
\end{table*}
\setlength{\tabcolsep}{1.4pt}

\paragraph{Ablation Studies.}

To verify the contributions of different components in our Cosh-DiT, we conduct ablation studies and construct three variants via component removal, which are described as follows:
\textbf{1)} w/o Hybrid Motion, we use raw 2D coordinates without 3D hand guidance.
\textbf{2)} w/o Spatial Alignment, we directly combine 2D skeletons with 3D hand meshes in the image plane without geometric-aware alignment. 
\textbf{3)} w/o Appearance Motion (AM)-DiT, we use single DiT blocks instead of iterative AM-DiT blocks.

Quantitative and qualitative studies are summarized in Tab.~\ref{tab:ablation} and Fig.~\ref{fig:ablation}. Both w/o Hybrid Motion and w/o Spatial Alignment produce undesirable artifacts in the hand regions, where the former causes hand texture degradation due to insufficient geometric prior utilization, while the latter causes severe misalignment between wrists and hands due to the lack of geometric-aware alignment process. The obvious drops in the confidence scores of detected key points in Tab.~\ref{tab:ablation} also support this observation.
w/o AM-DiT fails to preserve identity information well with only limited modeling ability from a single DiT block.
Our full model achieves the best performance among all the variants, which demonstrates the effectiveness of these components.

\begin{figure}[t]
    \centering
    \includegraphics[width=1.0\linewidth]{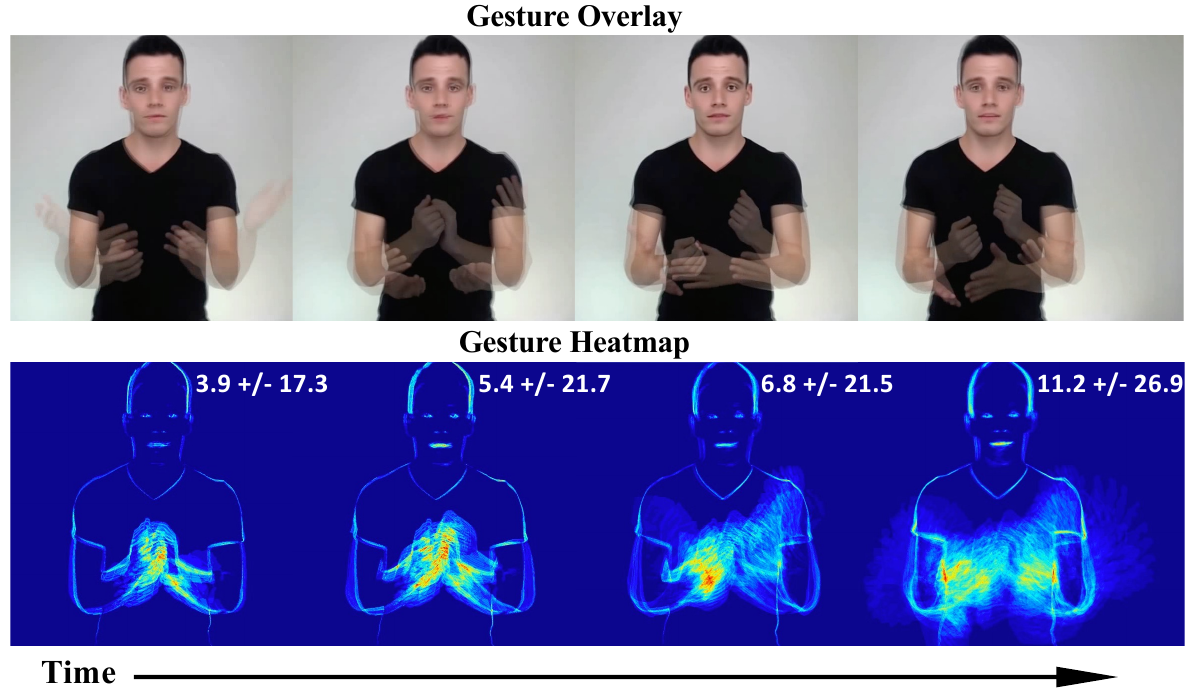}
    \caption{\textbf{Visual Analysis of Motion Diversity.} The top row shows an overlay of three generated videos, allowing for a comparative view of their motion patterns. In the bottom row, we present the accumulated motion overlay, highlighting the overall movement trends. The values in the top-right corner represent the scaled mean and standard deviation of the motion, offering a quantitative measure of the variability in the generated movements.
}
\label{fig:heatmap}

\vspace{-10pt}
\end{figure}

\paragraph{Motion Diversity.}
Given the many-to-many nature of this task, we provide an intuitive visualization of motion diversity in Fig.~\ref{fig:heatmap}.  
Specifically, we synthesize three videos using the same reference and human speech.  
To illustrate motion variations, we overlay the synthesized videos and generate an accumulated motion heatmap, scaled within $[0, 255]$.  
The upper-right corner presents the mean and standard deviation values over time.  
The varying hand gestures highlight the motion diversity captured by our approach, while the static background confirms that our system effectively isolates relevant regions for audio-to-motion mapping.

\paragraph{User Study.} 
Subjective evaluation is pivotal for assessing the quality of such systems. 
To this end, we conducted a user study with 60 participants who were asked to score the synthesized results based on the following criteria:

\begin{itemize}
    \item \textbf{Video Realness}: Evaluates image quality and temporal consistency.
    \item \textbf{Motion Naturalness}: Considers the naturalness of the synthesized motions and their alignment with human speech.
    \item \textbf{Identity Similarity}: Assesses appearance similarity, including facial features and attire.
\end{itemize}

Participants rated each video on a scale of 1 to 5. The average scores, representing the mean opinion, are reported in Tab.~\ref{table:MOS}.
The results demonstrate that our system outperforms competing approaches, producing more realistic video synthesis with natural body movements and superior alignment with the evaluation criteria.

\begin{figure}[t]
    \centering
    \includegraphics[width=1.0\linewidth]{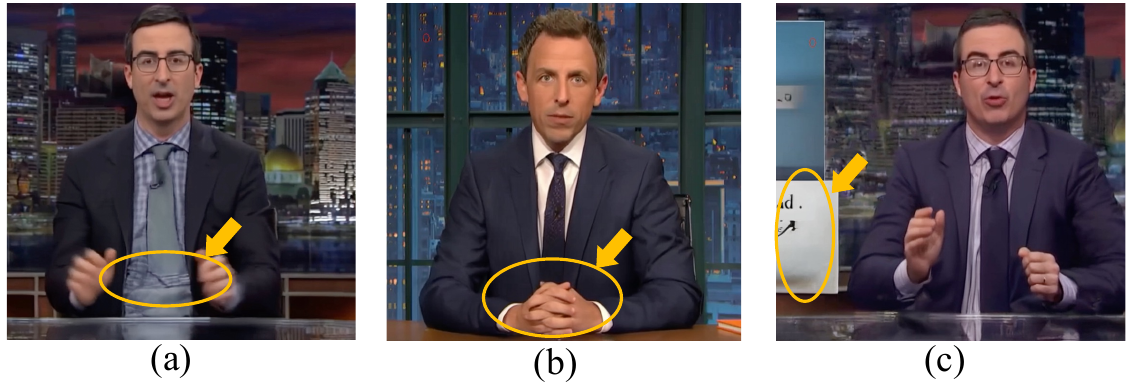}
    \caption{\textbf{Failure Cases.} Sometimes our model produces unsatifactory results at ties, overlapping hands and background synthesis when the reference image does not provide sufficient texture guidance.
}
\label{fig:failure}

\end{figure}

\paragraph{Failure Cases.} 
We highlight several cases where our approach faces challenges in Fig.~\ref{fig:failure}. Due to missing regions in the reference image, such as the ties, the synthesis is sometimes imperfect, as shown in Fig.~\ref{fig:failure}(a). For more complex hand gestures, the results still fall short of expectations, with unnatural shapes, as seen in Fig.~\ref{fig:failure}(b). Additionally, for fine-grained background details, such as text in Fig.~\ref{fig:failure}(c), our model still exhibits some artifacts.

\section{Conclusion}
In this work, we propose Co-Speech Gesture Video Synthesis via Hybrid
Audio-Visual Diffusion Transformers \textbf{Cosh-DiT}, a two stage framework that produces high-fidelity co-speech gesture videos through a discrete DiT and a continuous one. We identify the key aspects of our work. \textbf{1)} We explore audio-to-gesture generation on a different form of representation that is suitable for the human video synthesis task. It is nicely handled with a discrete diffusion transformer. This paradigm has never been explored before. \textbf{2)} We delicately designed our video-domain Diffusion Transformer model to take multiple inputs, including 3D renderings, previous motion frames and the reference frame, so that realistic co-speech gesture results can be produced. Our method clearer boosts the performance of co-speech gesture video generation.

\paragraph{Limitation.} The performance of our model is hindered by unstable backgrounds. Moreover, our model cannot handle results with cross fingers, which might cause difficulties in the 3D representations. Extending the capabilities of co-speech gesture video generation to diverse, real-world scenes remains a challenging open problem. Larger-scale pre-trained DiT models might be able to tackle these difficulties.

\paragraph{Ethical Consideration.} Our method has the potential to generate fabricated talks, raising concerns about potential misuse. To mitigate this risk, we are committed to strictly controlling the distribution of our models and the generated content, limiting access to research purposes only.

\paragraph{Data Availability Statement.} Due to privacy concerns, we are unable to release the dataset at this time. However, we will open-source it once we obtain consent for public release from the recorded participants.

{
    \bibliographystyle{sn-basic}
    \bibliography{sn-bibliography}
}

\end{document}